%% file: main.tex
\documentclass[]{fairmeta}
\usepackage{amsmath,amssymb}
\usepackage{array}
\usepackage{float}
\usepackage{makecell}
\usepackage{fontawesome5}
\hypersetup{colorlinks=true, linkcolor=red, citecolor=cyan, filecolor=magenta, urlcolor=magenta}
\title{Pixel is All You Need\IfFileExists{figs/shiba.png}{\hspace{0.3em}\raisebox{-0.18\height}{\includegraphics[height=1.15em]{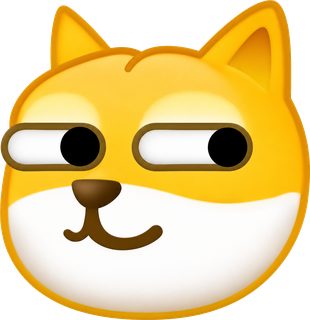}}}{}}

\author[1,2]{Jie Chu}
\author[1]{Haoting Shi}
\author[2]{Fucheng Zhang}
\author[1,2]{Jiawei Liu}
\author[1,2]{Junwei Liao}
\author[1,2]{Ying Wen}
\author[1]{Muning Wen}
\affiliation[1]{Shanghai Jiao Tong University}
\affiliation[2]{Shanghai Innovation Institute}

\abstract{Text, images, audio, touch, and even robot actions can, in principle, all be rendered onto a single canvas, suggesting an ambitious goal of unifying all modalities within vision. Language provides the natural first test. To subsume it, a visual model must reason directly over text in pixels. Across a wide range of image, video, and unified multimodal generators, however, we find that their visual generation modules cannot perform symbolic reasoning over text on the canvas. They succeed only when an underlying text module or prompt rewriter already supplies the answer, effectively reducing the generator to a mere renderer, a limitation that post-training does little to resolve. To test whether this failure stems from pixels themselves or from current training paradigms, we train a family of purely pixel-space generative models from scratch without any text pathway. With images as their sole input and output, these models read questions, write intermediate steps, and produce answers directly on the canvas. Remarkably, they exhibit hallmark behaviors of large language models. Performance improves predictably with scale following a power law. Writing intermediate steps frame by frame substantially outperforms single-frame answering, echoing chain-of-thought, while reinforcement-learning post-training further strengthens reasoning. Moreover, sharing one canvas allows the same model to reason across modalities, a capability that text-only language models lack. To the best of our knowledge, this is the first systematic study of these phenomena in purely pixel-space generative models free of text tokens.}

\date{\today}
\metadata[{\faGlobe\ Homepage}]{\url{https://bxzzcj.github.io/pixel-is-all-you-need}}
\metadata[{\faGithub\ Github}]{\url{https://github.com/bxzzcj/pixel-is-all-you-need}}

\begin{document}

\maketitle

\input{sec/1_intro}

\input{sec/2_related}
\input{sec/3_evaluation}
\input{sec/4_models}
\input{sec/5_scaling}

\input{sec/6_cot}
\input{sec/7_rl}
\input{sec/8_conclusion}
\input{sec/9_ack}
\input{sec/9_llm}

\bibliographystyle{assets/plainnat}
\bibliography{main}

\input{sec/X_suppl}

\end{document}

%% file: sec/1_intro.tex
\section{Introduction}
\label{sec:1}

\begin{figure}[t]
\centering
\includegraphics[width=1.0\textwidth]{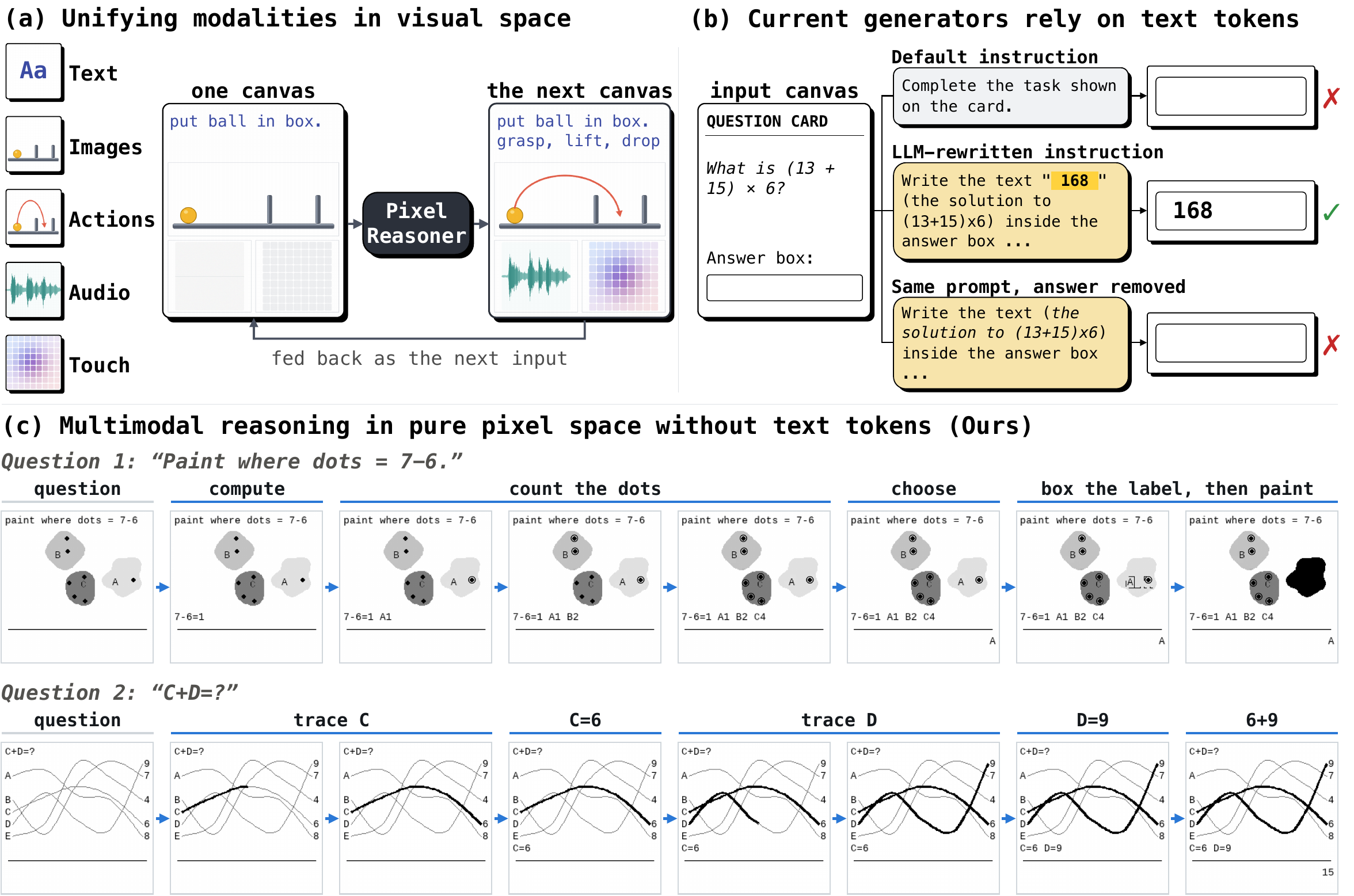}
\caption{Unifying modalities and reasoning in pixel space. (a) In our envisioned unified paradigm, text, images, audio, touch, and actions share one canvas, where a pixel-only model reads the current state, generates the next canvas, and feeds it back as input. (b) Current generators rely on text tokens. Such a generator typically writes the answer only when explicitly supplied with it, as shown by real outputs from Qwen-Image-2.1. (c) Trained from scratch without text tokens, our pixel model reasons across modalities frame by frame. Real rollouts from a 33M model on unseen problems interleave arithmetic, dot counting, and region painting, or curve tracing and summation.}
\label{fig:f1}
\end{figure}

Almost everything a machine needs to perceive or produce can be drawn. Text is a rendered string, an image is already a canvas, audio becomes a spectrogram, touch a pressure map, and a robot action a trajectory plotted in the workspace. If a single model could read and write all of these on one canvas, the boundaries between modalities would disappear, requiring no per-modality tokenizer, no cross-encoder adapter, and no separate decoder for each output type. A math problem and its solution, a chart and the numbers extracted from it, or a map and a route traced over it would all live in the same image, allowing the model to solve problems by updating that image step by step. This suggests an ambitious goal: \emph{unifying all modalities within vision} (\Cref{fig:f1}a).

Language is the natural first test of this goal, and the hardest one. Most symbolic reasoning, such as arithmetic, logic, and recalling or combining facts, takes place in language, which also serves as the reasoning medium for large language models (LLMs). For vision to subsume language, rendering visually plausible text is not enough. A visual model must reason directly over text on the canvas by reading a question written in pixels, solving it, and writing the answer back in pixels. Recent generators make this appear within reach. They render long passages of legible text~\citep{chen2023textdiffuser,yang2023glyphcontrol,wu2025qwenimage}, follow detailed editing instructions~\citep{brooks2023instructpix2pix,labs2025kontext}, and unify understanding and generation within a single network~\citep{deng2025bagel,zhou2024transfusion,xie2024showo,chen2025janus,team2024chameleon}, while video models have even been described as zero-shot reasoners~\citep{wiedemer2025video}.

However, we find that the generation modules of these systems cannot reason over text on the canvas. We construct a benchmark where questions appear in pixels and models write answers on the same canvas. Across image, video, and unified multimodal models, pixel-only questions, from word-level to arithmetic and factual tasks, are almost never answered correctly. Each of these models can be separated into a text module and a generation module, and image and video models are often preceded by an auxiliary prompt rewriter that reads the instruction and image to produce a detailed prompt for the text module. Our ablations show that these rewriters and text modules can usually solve the questions themselves, whereas generators succeed only when the rewriter or text module has already supplied the answer in text tokens (\Cref{fig:f1}b). Post-training teaches generation modules to copy and rearrange visible characters, yet arithmetic and factual accuracy remains near zero even on training items. Reasoning thus occurs in text tokens, and the visual generator merely renders the outcome of the prompt rewriter or text module. This echoes a common view that visual generators possess little intelligence and act largely as renderers for language models~\citep{guo2025mmecof,kou2026think,ren2026unireasoner}.

Is this a limitation of pixels themselves, or of how existing models are trained? Current generators are trained to \emph{render} content described in a text prompt, rather than to \emph{reason over} text embedded in their visual input~\citep{betker2023dalle3,chen2023textdiffuser,wu2025qwenimage,kou2026think}. To disentangle these two factors, we train a family of purely pixel-space generative models from scratch without any text pathway. Implemented as flow-matching transformers operating directly on raw pixels~\citep{li2025jit,lipman2023flow,peebles2023dit}, these models have no text tokens, no tokenizer, and no text encoder, so they take only pixels as input and produce only pixels as output. Every task is posed visually on a canvas. The question is printed on the canvas, the model writes its final answer on the same canvas, and when reasoning step by step, it generates a sequence of frames that progressively record intermediate steps onto the canvas (\Cref{fig:f1}c). We train models ranging from 7.6M to 458M parameters on arithmetic, multi-digit multiplication, word problems, factual question answering, and cross-modal tasks mixing text and pictures, and evaluate them on held-out problems never seen during training.

If pixels were the obstacle, these models would fail as well. They do not. Trained from scratch in this purely visual setting, they learn to read, reason, and write directly on the canvas, exhibiting three hallmark behaviors of LLMs, namely scaling laws, chain of thought, and effective reinforcement-learning (RL) post-training.

\textbf{Scaling laws.} Like LLMs~\citep{hoffmann2022chinchilla,kaplan2020scaling}, our pixel models improve predictably with scale. Crucially, this law extrapolates accurately. On text tasks, the per-step error falls from 0.174 at 7.6M parameters to 0.037 at 458M along a power law, and the law fitted only up to 130M predicts the held-out 458M error within 2\%.

\textbf{Chain of thought.} Writing intermediate steps frame by frame substantially outperforms single-frame direct answering, mirroring chain-of-thought reasoning in LLMs~\citep{wei2022cot}. At 130M with equal training, frame-by-frame reasoning solves 98.1\% of Compute Then Paint and 49.2\% of Trace Then Sum problems against 29.2\% and 17.0\% for direct answering, which reaches only 31.1\% and 17.2\% even when trained four times longer. On Word Problems, this advantage emerges with scale as accuracy climbs from 3.7\% at 7.6M to 78.6\% at 130M, while direct answering stays near 1\%, resembling the emergent abilities of LLMs~\citep{wei2022emergent}.

\textbf{Reinforcement learning.} RL post-training with verifiable rewards further strengthens visual reasoning~\citep{ouyang2022instructgpt,shao2024deepseekmath}. Learning from its own rollouts, the model discovers better reasoning trajectories. On Word Problems at 130M, GRPO raises accuracy from 23.7\% to 84.6\%, whereas supervised fine-tuning with the same compute reaches at most 39.0\%.

Beyond reproducing LLM-like behaviors, this pixel-only paradigm offers a capability that text-only LLMs inherently lack. Because questions, intermediate steps, and answers share a single canvas, one model can reason seamlessly across modalities. In \emph{Count Then Paint}, the model counts dots inside irregular regions, records the counts and target label in text, and paints the region with the most dots. In \emph{Trace Then Sum}, it traces intersecting curves and sums the values along them, and in \emph{Sum Mazes}, it draws a shortest path through a maze while tallying the numbers encountered along the route. All three LLM-like behaviors, including scaling, chain of thought, and RL gains, continue to hold on these cross-modal tasks.

Taken together, these findings demonstrate that the failure of current visual generators on direct textual and cross-modal reasoning stems not from pixels, but from training paradigms. In a purely visual world, a generative model can learn to read, compute, and write directly in pixels, and it exhibits the same foundational dynamics observed in LLMs, namely predictable scaling, gains from step-by-step intermediate computation, and substantial improvements under RL. Vision is therefore a credible substrate for unifying language with other modalities. To the best of our knowledge, this is the first systematic study of these phenomena in purely pixel-space generative models free of text tokens.

%% file: sec/2_related.tex
\section{Related Work}
\label{sec:2}

\textbf{Textual reasoning in visual generators.} Image editing and video models now render legible text~\citep{labs2025kontext,wan2025wan,wu2025qwenimage} and follow detailed instructions~\citep{labs2025kontext,brooks2023instructpix2pix}, yet benchmarks for reasoning-informed generation~\citep{niu2025wise,sun2025t2ireasonbench,wu2025krisbench,zhao2025risebench} reveal that they struggle on knowledge and reasoning. A common remedy places an LLM prompt rewriter before the visual generator~\citep{fang2025got,kou2026think,ren2026unireasoner}, leaving reasoning in text tokens. Other works process text as rendered images. PIXEL~\citep{rust2023pixel} and CLIPPO~\citep{tschannen2023clippo} learn language understanding from pixels, while DeepSeek-OCR~\citep{wei2025deepseekocr} compresses text contexts into vision tokens. Closest to us, PIXAR~\citep{tai2024pixar} and PixelGPT~\citep{chai2024pixelgpt} predict rendered text autoregressively as image patches. Trained at small scale on rendered corpora, they remain language models that output only text, using pixels merely as an input or output format. In our work, pixels are the medium of reasoning itself. A single model generates the question, intermediate steps, and answer across modalities on one canvas, allowing us to diagnose current generators and study scaling laws, chain of thought, reinforcement learning, and cross-modal reasoning.

\textbf{Pixel reasoning in visual generators.} Video generators exhibit chain-of-frames reasoning, solving mazes and visual puzzles zero-shot across frames~\citep{wiedemer2025video}, while Sora-2 has been prompted to solve math problems with a prompt rewriter~\citep{tong2025thinkingvideo}. However, a systematic evaluation of Veo-3 over 12 spatial, geometric, physical, temporal, and embodied dimensions shows that while such models handle short-horizon spatial coherence and local dynamics, they fail on long-horizon causal reasoning, strict geometry, and abstract logic, acting more as visual renderers paired with text models~\citep{guo2025mmecof}. Other approaches call visual generators to draw intermediate sketches~\citep{hu2024sketchpad} or images~\citep{li2025mvot} that assist multimodal LLMs. Because these studies evaluate pretrained video models zero-shot or use generated images merely as aids for a text model, the observed reasoning is either shallow or still resides in a text model. In contrast, we train pixel models from scratch with no text pathway. Every reasoning step is written directly in pixels by the generator itself, spanning expressions, word problems, factual questions, and summing values read from curves and mazes.

\textbf{Unified multimodal models.} Unified multimodal models perform understanding and generation in one network, either using a single transformer over text and image tokens~\citep{team2024chameleon,zhou2024transfusion,xie2024showo} or employing separate encoders or experts for understanding and generation~\citep{chen2025janus,deng2025bagel}. Many such systems can think in text tokens before generating an image, and this thinking step drives their gains on questions requiring knowledge or reasoning~\citep{deng2025bagel}. Our evaluation shows that the generation pathway of the unified models we test does not itself reason correctly over text on the canvas, since the generation module succeeds almost only when the text-token thoughts already contain the answer. Crucially, these architectures unify modalities in token space, where text remains a sequence of text tokens, images become tokens or latents in the same sequence or network, and reasoning happens in text tokens. We instead unify modalities in visual space itself. Text appears as pixels on the canvas, and a single pixel model without any text tokens reads, reasons, and writes across modalities on one shared canvas.

%% file: sec/3_evaluation.tex
\section{Evaluating the Textual Reasoning of Current Generators}
\label{sec:3}

We first examine whether existing visual generators can reason over text written on the canvas and, when they appear to succeed, where the reasoning actually takes place.

\subsection{Benchmark Design}
\label{sec:3.1}

\begin{figure}[t]
\centering
\includegraphics[width=0.62\linewidth]{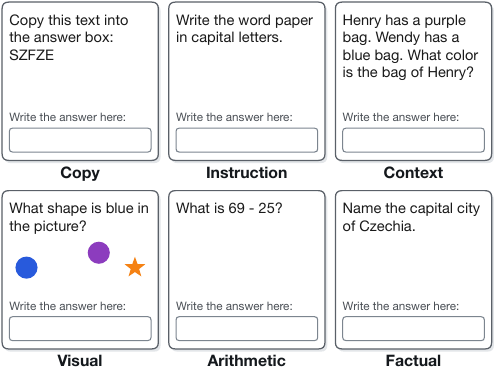}
\caption{The six tasks of the benchmark. Ordered by decreasing answer visibility from string copying to factual recall, each task tests whether a generator can read a question on the canvas and write the answer into the empty box.}
\label{fig:f2}
\end{figure}

Each item is a 1024\ensuremath{\times}1024 canvas containing a title, a printed question, a scene of colored shapes for the \emph{Visual} task, and an empty answer box (\Cref{fig:f2}). In the default condition, the model receives only the canvas alongside a fixed text instruction that is identical across all items, namely ``Complete the task shown on the card. Keep everything else on the card unchanged''. Its text module may use its own tokenizer and text encoder to process this instruction. Six tasks span decreasing degrees of answer visibility on the canvas, meaning how much of the correct answer already appears in the question itself, as shown in \Cref{fig:f2}. Specifically, the benchmark comprises \emph{Copy} (copying a string, 200 test items), \emph{Instruction} (applying an instruction to a word, 216), \emph{Context} (selecting a word from a short passage, 300), \emph{Visual} (answering a question about a scene, 400), \emph{Arithmetic} (300), and \emph{Factual} (answering a factual question, 579). We employ Qwen3.8-Max, a strong multimodal model, as the judge to determine whether the output of a generator is correct.

Four diagnostic conditions control whether and in what form the generation module sees the answer. For image editing and video models, Rewrite passes the instruction and image through a prompt rewriter that produces a detailed prompt often containing the answer, and Delete removes the answer from that rewritten prompt. For unified multimodal models, Render writes the answer into the prompt as text tokens, such as instructing the model to write ``Madrid'', the capital of Spain, in the answer box, and Think lets the text module reason in text tokens before the image is generated. Appendix~\ref{app:A} gives the details.

\subsection{Results on Current Generators}
\label{sec:3.2}

As shown in \Cref{tab:t1}, current visual generators answer almost nothing in the default condition. Open image editing models, including FLUX.3-klein, Qwen-Image-2.1, and LongCat-Image-Edit, score 0\% on every task, even on \emph{Copy} where the answer is already visible in the question. Video generation models fail in the same way, leaving the answer box empty in most generated frames and often answering no item at all. Unified multimodal models likewise answer almost nothing when generating directly, with BAGEL, InternVL-U, and Emu3.5 scoring 0\% and SenseNova reaching only 0\% to 5\% outside the \emph{Copy} task. By contrast, closed systems such as GPT-Image-2.5, Gemini-3-Pro-Image, and Nano-Banana-2.1 answer most items, with averages of 98.0\%, 94.1\%, and 92.0\%. However, our experiments in \Cref{tab:t1} and Appendix~\ref{app:B}, together with public analyses of these systems, indicate that their success relies essentially on an external or built-in text reasoning step. OpenAI documents that in its image generation tool the LLM revises the prompt before image generation~\citep{openai2026imagegen}, Gemini image models always think in text tokens before drawing~\citep{google2026geminithinking}, and a diagnostic study of GPT-4o image generation points to a language model followed by an image decoder~\citep{yan2025gptimgeval}.

\begin{table}[t]
\centering
\small\setlength{\tabcolsep}{6pt}
\begin{tabular}{@{}lccccc@{}}
\toprule
\textbf{Model} & \textbf{Default} & \textbf{Rewrite} & \textbf{Delete} & \textbf{Render} & \textbf{Think} \\
\midrule
\multicolumn{6}{l}{\textit{Image editing models}} \\
GPT-Image-2.5 & -- & \textbf{98.0} & -- & -- & -- \\
Gemini-3-Pro-Image & -- & \textbf{94.1} & -- & -- & -- \\
Nano-Banana-2.1 & -- & \textbf{92.0} & -- & -- & -- \\
Qwen-Image-2.0-Pro & 0.0 & \textbf{91.7} & -- & -- & -- \\
Qwen-Image-2.1 & 0.0 & \textbf{86.0} & 0.0 & -- & -- \\
FLUX.3-klein & 0.0 & \textbf{83.4} & 0.0 & -- & -- \\
Wan2.6-Image & 0.0 & \textbf{81.7} & -- & -- & -- \\
FLUX.2-klein & 0.0 & \textbf{55.5} & 0.0 & -- & -- \\
LongCat-Image-Edit & 0.0 & \textbf{20.4} & 0.0 & -- & -- \\
\midrule
\multicolumn{6}{l}{\textit{Video generation models}} \\
MiniMax-H3 & 1.4 & \textbf{98.3} & 0.0 & -- & -- \\
Wan 3.0 & 0.0 & \textbf{83.3} & -- & -- & -- \\
OmniWeaving & 0.0 & \textbf{24.0} & 0.0 & -- & -- \\
Cosmos3 & 0.0 & \textbf{13.6} & 0.0 & -- & -- \\
Wan2.2 & 0.0 & \textbf{0.7} & 0.0 & -- & -- \\
\midrule
\multicolumn{6}{l}{\textit{Unified multimodal models}} \\
SenseNova & 18.7 & -- & -- & \textbf{99.7} & 96.0 \\
Emu3.5 & 0.0 & -- & -- & \textbf{27.0} & 0.5 \\
BAGEL & 0.0 & -- & -- & \textbf{13.3} & 0.7 \\
InternVL-U & 0.0 & -- & -- & 1.3 & \textbf{12.0} \\
\bottomrule
\end{tabular}
\caption{Exact accuracy in percent of current visual generators. In \textbf{Default}, models receive only the canvas and a fixed prompt without the answer. For image and video models, \textbf{Rewrite} uses a prompt rewriter that often supplies the answer, and \textbf{Delete} removes the answer from the rewritten prompt. For unified models, \textbf{Render} writes the answer into the prompt as text tokens, and \textbf{Think} turns on thinking mode, enabling text-token reasoning before generation. Video models are judged on the final frame. Dashes mark inapplicable or unavailable conditions, as closed systems whose prompt rewriter cannot be disabled lack Default, and systems whose internal rewritten prompt cannot be edited lack Delete. The highest value per row is in bold.}
\label{tab:t1}
\end{table}

\subsection{Where the Reasoning Happens}
\label{sec:3.3}

\begin{figure}[t]
\centering
\includegraphics[width=0.62\linewidth]{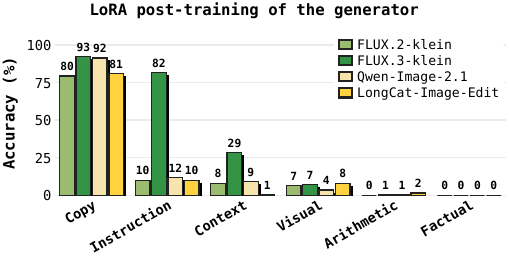}
\caption{Test-set accuracy after post-training the generation module of four image editing models. All base models start at 0\% on every task. Post-training teaches the generators to copy visible strings, yet accuracy falls steeply as answer visibility in the question decreases and stays at 0\% to 2\% on Arithmetic and Factual.}
\label{fig:f3}
\end{figure}

We find that the only path to a correct answer on the canvas is for that answer to be first produced as text tokens by the prompt rewriter of an image editing or video model, or by the text module of a unified multimodal model. As shown in \Cref{tab:t1}, after a rewriter expands the prompt, image editing and video models draw the answer correctly in most cases because the rewritten prompt often contains the answer. For example, Qwen-Image-2.1 reaches 86.0\% with a rewriter, and MiniMax-H3 reaches 98.3\%. Yet when only the answer is deleted from the rewritten prompt while everything else is kept, the success rate of every tested generator drops to near 0\%. The rewriter thus helps by directly supplying the answer to the generation module.

The text modules themselves possess the reasoning capability. Used as chat models, the text encoders of FLUX.2-klein, FLUX.3-klein, Qwen-Image-2.1, and LongCat-Image-Edit answer 86\% to 97\% of Factual items and 83\% to 100\% of Arithmetic items in text tokens. When the conditioning states passed from the text encoder to the generator in Qwen-Image-2.1 are decoded with the encoder's own language-model head, the first token of the correct answer ranks among the top 10 for 140 of 150 Factual questions, yet the generator writes none of these answers, as detailed in Appendix~\ref{app:D}.

\begin{figure}[t]
\centering
\includegraphics[width=1.0\textwidth]{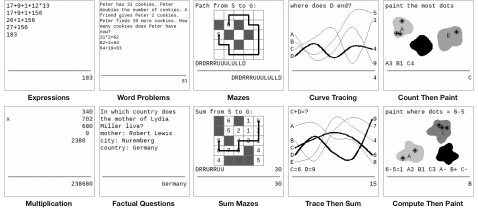}
\caption{Overview of the ten pixel-space reasoning tasks. Each 192\ensuremath{\times}192 (256\ensuremath{\times}256 for Word Problems) canvas places the question and correct intermediate reasoning steps above a horizontal rule, with the final answer written below.}
\label{fig:f4}
\end{figure}

Unified multimodal models likewise reason only in text tokens. SenseNova, which thinks in text tokens before generating, answers 96.0\% of items in thinking mode (\Cref{tab:t1}). Reading its thoughts item by item reveals that the generated image is correct on 99.0\% of the items whose thought contains the answer, compared to only 10\% whose thought does not. InternVL-U follows the same pattern, as Appendix~\ref{app:D} shows.

\subsection{Post-Training the Generation Module}
\label{sec:3.4}

We next test whether post-training the generation module enables it to reason directly on the canvas. For FLUX.2-klein, FLUX.3-klein, Qwen-Image-2.1, and LongCat-Image-Edit, we train one LoRA per task that maps a canvas with an empty box to the same canvas with the answer written in, as detailed in Appendix~\ref{app:C}. As shown in \Cref{fig:f3}, accuracy falls steeply as answer visibility decreases. It reaches 80\% to 93\% on \emph{Copy}, where the answer appears in the question, and only 10\% to 12\% on new words in \emph{Instruction}, where FLUX.3-klein, at 82\%, masters only the five trained rules but scores 0\% on an unseen one, as \Cref{tab:t5} in Appendix~\ref{app:C} shows. It falls further to at most 29\% on \emph{Context}, below 10\% on \emph{Visual}, and 0\% to 2\% on \emph{Arithmetic} and \emph{Factual}. These accuracies hold on training and test sets alike, so the models learn to copy visible text rather than to think or recall.

Fine-tuning the generation module of a unified multimodal model yields the same result. BAGEL's text module answers 83\% of 450 benchmark items in text tokens, yet training its generation module, alone or together with the text module, to write the answer keeps Default accuracy at 0\% to 1.8\%, while the same training lifts Render from 47\% to 60\% on these items, as detailed in Appendix~\ref{app:C}.

\subsection{Summary of Investigation}
\label{sec:3.5}

Across all generation systems whose pipelines we can inspect, nearly every correct answer on the canvas is preceded by the same answer in text tokens. Their generation modules can read the canvas and write text, yet reasoning occurs in the prompt rewriter or text module, and the generators only render the result~\citep{kou2026think,ren2026unireasoner}. Whether a pixel generator trained from scratch without any text pathway can learn to reason over text on the canvas is the question we turn to next.

%% file: sec/4_models.tex
\section{Pixel-Space Generative Models without Text Tokens}
\label{sec:4}

We now remove the text pathway entirely and construct a purely visual world. This section introduces the tasks, model architecture, and training regimes.

\subsection{Tasks}
\label{sec:4.1}

Every problem is presented on a square grey canvas. Text is rendered in a fixed monospaced font on a grid of 24 columns and 12 rows of 8\ensuremath{\times}16-pixel cells. The question occupies the upper rows, a horizontal rule separates it from two answer rows at the bottom, and visual elements such as curves, regions, and dots are drawn in the remaining space. The model never receives character codes, seeing only pixels and writing only pixels. \Cref{fig:f4} illustrates all ten tasks. Four text tasks, namely \emph{Expressions}, \emph{Multiplication}, \emph{Word Problems}, and \emph{Factual Questions}, mirror the symbolic reasoning that LLMs perform in text. Six cross-modal tasks, namely \emph{Mazes}, \emph{Sum Mazes}, \emph{Curve Tracing}, \emph{Trace Then Sum}, \emph{Count Then Paint}, and \emph{Compute Then Paint}, interleave text and pictures in both the problem statement and the solution process.

\subsection{Model and Training}
\label{sec:4.2}

The model is a flow-matching transformer operating directly on pixels~\citep{lipman2023flow,liu2023rectified,peebles2023dit}, with no VAE, no tokenizer, and no text encoder. Following JiT~\citep{li2025jit}, the network predicts the clean image and is trained with the velocity loss. Its input is the channel-wise concatenation of the conditioning canvas (the original problem canvas, or the current canvas in frame-by-frame reasoning) and the noisy target canvas, partitioned into 576 patches of 8\ensuremath{\times}8 pixels, while the time step enters through adaptive layer normalization. We train five model sizes ranging from 7.6M to 458M parameters under a unified recipe, as detailed in \Cref{tab:t6} of Appendix~\ref{app:F}.

Using the same architecture, data, and training budget, we train the model under two regimes. \emph{Direct answering} maps the original problem canvas to the final canvas with the answer in a single generation, without any written intermediate steps. \emph{Frame-by-frame reasoning} trains the model on pairs of consecutive frames, mapping the canvas at step t to the canvas at step t+1. At test time, it generates frames sequentially starting from the original problem canvas, feeding each generated frame directly back in raw pixel space as the input for the next step. The canvas thus serves as the sole memory, so everything the model computes must be recorded on it. Rollout terminates once the model writes an answer into the answer box, allowing the model itself to decide how many frames to draw. In this setup, frame-by-frame reasoning corresponds to chain of thought, whereas direct answering corresponds to answering without intermediate thought. Further training and evaluation details are provided in Appendices~\ref{app:F} and~\ref{app:G}.

%% file: sec/5_scaling.tex
\section{Early Stage of Scaling Law}
\label{sec:5}

LLMs that reason in text tokens improve predictably with parameter count~\citep{kaplan2020scaling,hoffmann2022chinchilla}, as do diffusion transformers that render images in pixel space~\citep{liang2024scalingdit}. We examine whether reasoning performed directly in pixels obeys the same scaling law.

\subsection{Setup and Metric}
\label{sec:5.1}

\begin{figure}[t]
\centering
\includegraphics[width=0.56\linewidth]{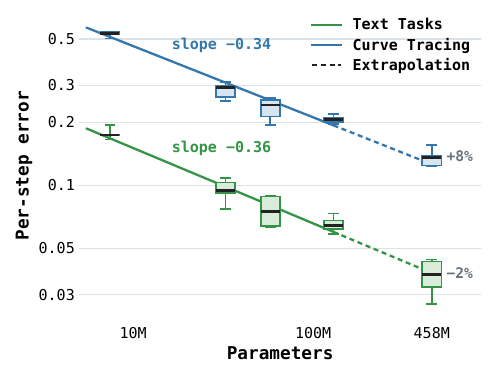}
\caption{Pixel-space reasoning obeys power-law scaling. From 7.6M to 458M parameters, per-step error in generating the next frame falls linearly on log-log scales for two models, one evaluated across Expressions, Multiplication, and Factual Questions and the other on Curve Tracing. Power laws fitted to medians from 7.6M to 130M extrapolate accurately to the held-out 458M model, with dashed lines matching measured medians within 2\% and 8\%. Across five seeds per size, boxes cover the middle three, whiskers span all five, and black bars mark medians.}
\label{fig:f5}
\end{figure}

We train frame-by-frame models at five scales from 7.6M to 458M parameters with one recipe and five seeds per size. We measure the per-step error, the visual analogue of per-token error in an LLM. It is defined as the fraction of next frames generated from a correct canvas that contain a wrong character on text tasks or whose changed pixels have an IoU below 0.8 with the ground-truth update on cross-modal tasks. We train two joint models and fit a power law $L(N) = A\,N^{-\alpha}$ to each resulting per-step error curve. The first is trained on Expressions with four steps, Multiplication with 4\ensuremath{\times}4 digits, and Factual Questions and evaluated on their average. The second is trained on Mazes, Curve Tracing, and Count Then Paint, and since Mazes and Count Then Paint saturate from 33M on, it is evaluated on Curve Tracing.

\subsection{Power-Law Fit}
\label{sec:5.2}

\begin{figure}[t]
\centering
\begin{minipage}[t]{1.0\textwidth}\vspace{0pt}{\small\textbf{(a)}}\\[1pt]\includegraphics[width=\linewidth]{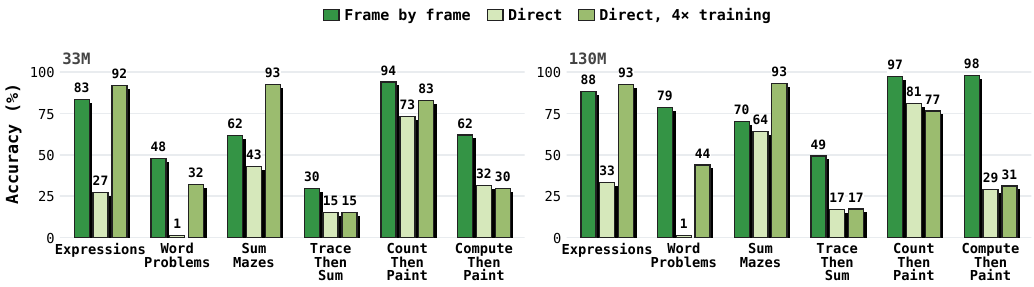}\end{minipage}\\[0pt]
\begin{minipage}[t]{1.0\textwidth}\vspace{0pt}{\small\textbf{(b)}}\\[1pt]\includegraphics[width=\linewidth]{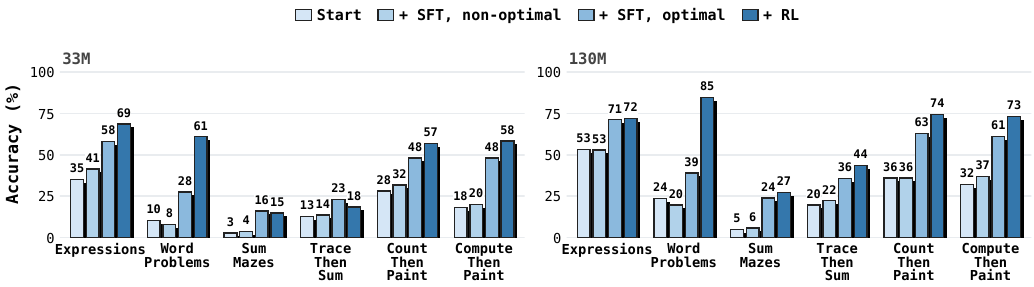}\end{minipage}
\caption{Frame-by-frame chain of thought and reinforcement learning at 33M and 130M. (a) Writing steps frame by frame outperforms direct answering across all six tasks at equal training of 40k steps and surpasses four times longer direct training on four of them. (b) On the four cross-modal tasks, each run continues from a Start model trained on non-optimal demonstrations with extra steps or detours. Under equal compute, RL outperforms non-optimal SFT and surpasses optimal SFT on four tasks at 33M and all six at 130M. We apply GRPO to the two text tasks and RFT to the four cross-modal tasks. For the four multimodal tasks, test accuracy requires a correct answer in the fewest frames.}
\label{fig:f6}
\end{figure}

As shown in \Cref{fig:f5}, per-step error follows a power law on both curves and extrapolates accurately to a larger model. For each curve, we fit the power law using the median error across the five seeds at each scale from 7.6M to 130M and predict the performance of the held-out 458M model. On the text tasks, the median error falls from 0.174 at 7.6M to 0.064 at 130M, following L(N) \ensuremath{\propto} N\textsuperscript{\ensuremath{-}0.360}, and the measured error of the 458M model lies 2\% below the prediction. On \emph{Curve Tracing} in the joint cross-modal model, the error follows L(N) \ensuremath{\propto} N\textsuperscript{\ensuremath{-}0.340}, with the 458M model measuring 8\% above the prediction. \Cref{tab:t7} in Appendix~\ref{app:J} reports the fitted curves, the 90\% intervals for the predictions, measured medians, and relative errors, as well as the details of how they are computed.

%% file: sec/6_cot.tex
\section{Chain of Thought}
\label{sec:6}

In LLMs, generating intermediate reasoning steps before producing the final answer improves performance~\citep{nye2021scratchpad,wei2022cot}. Our pixel models can perform the same computation directly on the canvas. Using the same model architecture, training data, and training budget, we compare frame-by-frame reasoning against direct answering.

With an equal training budget of 40k steps, writing intermediate steps frame by frame substantially outperforms direct answering across all six tasks, as shown in \Cref{fig:f6}a. On \emph{Expressions} task, frame-by-frame reasoning solves 83.4\% and 88.3\% of unseen problems at 33M and 130M parameters, compared to 27.1\% and 33.2\% for direct answering. On \emph{Word Problems}, frame-by-frame accuracy reaches 47.8\% and 78.6\%, whereas direct answering remains at 1.1\% and 1.2\%. Training direct answering four times longer, for 160k steps, barely narrows the gap. At 33M and 130M parameters, frame-by-frame reasoning trained for only 40k steps achieves 29.5\% and 49.2\% on \emph{Trace Then Sum} and 61.9\% and 98.1\% on \emph{Compute Then Paint}, while direct answering trained for 160k steps reaches only 15.2\% and 17.2\% on \emph{Trace Then Sum} and 29.5\% and 31.1\% on \emph{Compute Then Paint}. Writing steps frame by frame is thus both far more accurate and much faster to learn.

Crucially, this advantage emerges with scale. On \emph{Word Problems}, frame-by-frame accuracy climbs from 3.7\% at 7.6M parameters to 47.8\% at 33M and 78.6\% at 130M, while direct answering stays near 1\% across all model sizes. Mirroring chain-of-thought behavior in LLMs~\citep{wei2022cot,wei2022emergent}, generating intermediate steps helps only once the model is large enough to execute each individual step reliably, as detailed in Appendix~\ref{app:I}.

%% file: sec/7_rl.tex
\section{Reinforcement-Learning Post-Training}
\label{sec:7}

Reinforcement learning with verifiable rewards improves the reasoning of LLMs beyond supervised training~\citep{ouyang2022instructgpt,shao2024deepseekmath,guo2025deepseekr1}. We investigate whether the same holds for reasoning conducted in pixels.

\subsection{Experimental Setup}
\label{sec:7.1}

On the two text tasks in \Cref{fig:f6}b, we apply GRPO~\citep{shao2024deepseekmath} to entire multi-frame rollouts and reward an exact answer read from pixels, as detailed in Appendix~\ref{app:H}. On the four cross-modal tasks, we use rejection-sampling fine-tuning, or RFT, which trains on the model's own successful rollouts.

To test whether RL discovers better solutions rather than imitating data, we continue from the same supervised start model with RL or with the same compute of supervised fine-tuning on optimal or non-optimal demonstrations, termed optimal SFT and non-optimal SFT. On the cross-modal tasks, the non-optimal demonstrations contain detours. In \emph{Sum Mazes}, they follow a longer path. In \emph{Count Then Paint} and \emph{Compute Then Paint}, half of the demonstrations first box an incorrect region and correct it in the next frame, while in \emph{Trace Then Sum}, half first trace an unrequested curve. Optimal demonstrations take the fewest frames and no detour. On the text tasks, the non-optimal demonstrations repeat steps, as detailed in Appendix~\ref{app:H}. The start model is trained on the non-optimal demonstrations for the cross-modal tasks and on optimal ones for the text tasks. Whenever RL outperforms optimal SFT, it shows that the model has found better solutions through its own rollouts.

\subsection{Comparison with Supervised Training}
\label{sec:7.2}

As shown in \Cref{fig:f6}b, RL improves reasoning across all six tasks. Non-optimal SFT yields little gain, whereas RL, learning solely from its own rollouts, matches or outperforms optimal SFT in most settings. At 130M parameters, RL surpasses optimal SFT on every task, reaching 84.6\% compared to 39.0\% on \emph{Word Problems}, 43.7\% compared to 35.7\% on \emph{Trace Then Sum}, and 73.2\% compared to 61.0\% on \emph{Compute Then Paint}. At 33M parameters, RL outperforms optimal SFT on \emph{Expressions}, \emph{Word Problems}, \emph{Count Then Paint}, and \emph{Compute Then Paint}.

Crucially, RL reshapes how the model reasons. On \emph{Expressions}, it learns to take smaller steps across more frames, trading a longer rollout for fewer per-frame errors, much as RL lengthens reasoning chains in LLMs~\citep{guo2025deepseekr1}. For example, on this task, the average rollout length grows from 3.34 to 4.53 frames at 33M and from 3.36 to 4.01 frames at 130M, while supervised training remains between 3.3 and 3.4 frames. On the cross-modal tasks, RL learns to eliminate the detours present in its initial demonstrations.

%% file: sec/8_conclusion.tex
\section{Conclusion}
\label{sec:8}

We asked whether vision can unify other modalities, starting with symbolic reasoning over text on a canvas. Across current image editing, video generation, and unified multimodal systems, visual generation modules cannot reason on their own and succeed only when an external prompt rewriter or built-in text module already supplies the answer in text tokens, a limitation post-training barely resolves. Purely pixel-space models trained from scratch without any text pathway instead learn to read, reason, and write on the canvas. Their per-step error on held-out problems follows a power law in parameter count. Frame-by-frame reasoning far outperforms direct answering with gains emerging at scale, and RL strengthens reasoning further. Because text and pictures share one canvas, a single model can also reason seamlessly across modalities, a capability text-only LLMs lack. Together, these findings indicate that the failure of current visual generators stems from their training rather than from pixels themselves, and establish vision as a credible substrate for unifying text and other modalities. Scaling this paradigm up is the natural next step.

%% file: sec/9_ack.tex
\section*{Acknowledgements}
We thank Wenhao Yan and Zihan Zhang for their deep insights into foundation models, and Wenjun Xiong, Can Chen, and Shengtao Zhang for the daily research discussions and their company throughout this work.

%% file: sec/9_llm.tex
\section*{LLM Usage Statement}
We used LLM-based assistants, Claude (Anthropic) and Codex (OpenAI), for editing and grammar of the text, for drafting and revising parts of the appendix from our experiment logs and code, for converting the manuscript to LaTeX, for writing plotting and figure-layout code, and for implementing and debugging experiment and evaluation code. The research questions, experimental design, analyses and conclusions are the authors' own. The authors checked every reported number against the experiment logs and take full responsibility for the content of the paper. LLMs are also part of the experimental setup of \Cref{sec:3}. The prompt rewriters we study are the official upsamplers and rewriting rules of each system, including the built-in rewriters of commercial systems. Open rewriters run with their own language models, and Qwen3.8-Max runs those whose official version calls a hosted model or is not released. Qwen3.8-Max also deletes the answers from rewritten prompts and serves as the judge, as described in Appendix~\ref{app:A}.

%% file: sec/X_suppl.tex
\clearpage
\beginappendix
\crefalias{section}{appendix}

\section{The benchmark of \texorpdfstring{\Cref{sec:3}}{Section 3}}
\label{app:A}

\textbf{Canvases.} Each item is a 1024\ensuremath{\times}1024 canvas with a printed title, a printed question, and an empty answer box under the label ``Write the answer here:''. Every model receives the canvas along with the identical default instruction ``Complete the task shown on the card. Keep everything else on the card unchanged.'' across all items. Video generation models receive the canvas as their first frame.

\textbf{Tasks and test sets.} 25\% of the source material of each task is held out for testing. \emph{Copy} asks for a random string over a 32-character alphabet to be copied, and is tested on new strings. \emph{Instruction} applies a rule to a word, such as writing it backwards, giving its first or last letter, writing it in upper case, or writing it twice. One test set applies trained rules to new words, and another applies a rule never trained, writing the word without its first letter. \emph{Context} asks for a word from a short passage involving named people, and is tested on trained passage types with new names. \emph{Visual} answers a question about a scene of colored shapes and is tested on new scenes. \emph{Arithmetic} evaluates a short expression and is tested on new expressions. \emph{Factual} asks questions over six relations, namely capital, continent, language, chemical symbol, author, and antonym, and is tested on paraphrases of trained facts, reversed questions, two-hop questions, and untrained facts.

\textbf{Judge.} Qwen3.8-Max transcribes the text inside the answer box, in whatever style it is written, without seeing the question or the answer key. An item is marked correct when the transcription matches the answer exactly after normalizing whitespace, and regardless of case on tasks that ignore case. For videos, the judge reads 8 evenly spaced frames, and we report both the final frame and whether any frame is correct.

\textbf{Conditions.} Beyond Default, four conditions control whether and in what form the generation module sees the answer. Rewrite expands the instruction and the input image into a more detailed, structured, and constraining prompt before the text module encodes it. Delete removes any answer from that rewritten prompt. Render writes the answer into the prompt as text tokens, for example asking the model to write ``Madrid'', the capital of Spain, in the answer box. Think lets the text module of a unified model first reason about the canvas in text tokens and then generate the image from that reasoning.

\textbf{Rewrites.} The image editing models use their official prompt upsamplers or rewriting rules. Wan2.2 and OmniWeaving share one rewrite, produced by Wan2.2's official image-to-video prompt extension run with Qwen3.8-Max, Cosmos3 uses its own JSON prompt upsampler, and MiniMax-H3 uses a rewrite that follows its official prompt guide. Under \emph{Delete}, Qwen3.8-Max removes every trace of the answer from that same rewrite, including accented variants, spellings written letter by letter, and trailing fragments such as ``, D.C.'' left from full answers like Washington, D.C. If a rewrite consists solely of the answer, it is replaced with the neutral phrase ``the answer''.

\begin{table}[!htb]
\centering
\small
\resizebox{\ifdim\width>\linewidth\linewidth\else\width\fi}{!}{%
\begin{tabular}{lccccccc}
\toprule
\textbf{System, condition} & \textbf{Factual (paraphrase)} & \textbf{Factual (unseen)} & \textbf{Arithmetic} & \textbf{Instruction} & \textbf{Context} & \textbf{Visual} & \textbf{Copy} \\
\midrule
\multicolumn{8}{l}{\textit{Image editing models, Rewrite (Default is 0 on every set for all six models)}} \\
FLUX.2-klein, official prompt upsampler & 57 & 57 & 73 & 53 & 45 & 60 & 43 \\
FLUX.3-klein, official prompt upsampler & 92 & 90 & 78 & 63 & 77 & 87 & 97 \\
Qwen-Image-2.1, API rewrite & 95 & 90 & 92 & 92 & 95 & 95 & 43 \\
LongCat-Image-Edit, upsampler-style rewrite & 37 & 45 & 25 & 5 & 17 & 12 & 2 \\
Qwen-Image-2.0-Pro, its own rewriting & 100 & -- & 90 & 100 & 100 & 100 & 60 \\
Wan2.6-Image, its own rewriting & 100 & -- & 90 & 90 & 80 & 100 & 30 \\
GPT-Image-2.5 & 95 & 98 & 100 & 95 & 100 & 98 & 100 \\
Gemini-3-Pro-Image & 93 & 92 & 98 & 88 & 95 & 93 & 100 \\
Nano-Banana-2.1 & 93 & 97 & 98 & 88 & 77 & 98 & 93 \\
\midrule
\multicolumn{8}{l}{\textit{Unified multimodal models}} \\
BAGEL, Default / Think & 0 / 0 & 0 / 0 & 0 / 0 & 0 / 1 & 0 / 0 & 0 / 0 & 0 / 4 \\
InternVL-U, Default / Think & 0 / 7 & 0 / 12 & 0 / 12 & -- & -- & 0 / 7 & 0 / 23 \\
SenseNova, Default / Think & 0 / 93 & 0 / 97 & 2 / 100 & -- & -- & 5 / 98 & 87 / 92 \\
Emu3.5, Default / Think & \multicolumn{7}{c}{0 on every set / 0.5 over all items} \\
\bottomrule
\end{tabular}}
\caption{\Cref{tab:t1} per test set, as exact accuracy in percent. Instruction holds new words under trained rules. With the answer deleted from the rewrite, all four open image editing models score 0 on Factual and Arithmetic. In SenseNova's Think mode, the image is correct on 99.0\% of the items whose thought contains the answer.}
\label{tab:t2}
\end{table}

\begin{table}[!htb]
\centering
\small
\resizebox{\ifdim\width>\linewidth\linewidth\else\width\fi}{!}{%
\begin{tabular}{lllll}
\toprule
\textbf{Video generation model} & \textbf{Default, final frame} & \textbf{Default, any of 8 frames} & \textbf{Rewrite} & \textbf{Delete} \\
\midrule
Wan2.2 & 0 & 0 & 0.7 & 0 \\
OmniWeaving & 0 & 0.2 & 24.0 (any frame: 28.8) & 0 \\
Cosmos3 & 0 & 0 & 13.6 (any frame: 17.4) & 0 \\
MiniMax-H3 & 1.4 & 1.4 & 98.3 & 0 \\
Wan 3.0 & 0 & 1.7 & 83.3 & -- \\
\bottomrule
\end{tabular}}
\caption{Video generation models, as exact accuracy in percent. Default gives the canvas as the first frame and a fixed prompt without the answer, judged on 8 evenly sampled frames. In 88\% to 98\% of frames from Wan2.2, OmniWeaving, and Cosmos3, the answer box remains empty. Delete is evaluated on Factual and Arithmetic, where the answer cannot be read off the canvas. For Wan 3.0 the rewrite is its own, with prompt\_extend on as described in Appendix~\ref{app:B}.}
\label{tab:t3}
\end{table}

\section{Prompt rewriting in commercial systems}
\label{app:B}

We call Qwen-Image-2.0-Pro, Wan2.6-Image, and Wan 3.0 through the official Alibaba Model Studio interface, whose parameter prompt\_extend turns the model's own prompt rewriting on or off. The image editing models receive the canvas as an image and the default instruction, with Wan2.6-Image in editing mode at 1K output. Wan 3.0 receives the canvas as its first frame and a fixed instruction to write only the answer in the answer box, generating 5 s of video at 480P or 720P. Each item uses a fixed random seed that stays identical whether the rewriter is off or on, and generation is reproducible within 0.55 grey levels. Toggling the rewriter switch alone therefore raises accuracy from 0\% to as much as 91.7\%, as \Cref{tab:t4} shows. The items are drawn with a fixed seed from six test sets, namely Copy, Instruction, Context, Visual, Arithmetic, and Factual paraphrases. Because GPT-Image-2.5, Gemini-3-Pro-Image, and Nano-Banana-2.1 always run their language model, \Cref{tab:t2} reports them with rewriting enabled.

\begin{table}[!htb]
\centering
\small
\resizebox{\ifdim\width>\linewidth\linewidth\else\width\fi}{!}{%
\begin{tabular}{llccccccc}
\toprule
\textbf{System (official interface)} & \textbf{Rewriter} & \textbf{All tasks} & \textbf{Copy} & \textbf{Instruction} & \textbf{Context} & \textbf{Visual} & \textbf{Arithmetic} & \textbf{Factual} \\
\midrule
\multirow{2}{*}{Qwen-Image-2.0-Pro (image editing)} & off & 0 & 0 & 0 & 0 & 0 & 0 & 0 \\
 & on & 91.7 & 60 & 100 & 100 & 100 & 90 & 100 \\
\multirow{2}{*}{Wan2.6-Image (image editing)} & off & 0 & 0 & 0 & 0 & 0 & 0 & 0 \\
 & on & 81.7 & 30 & 90 & 80 & 100 & 90 & 100 \\
\multirow{2}{*}{Wan 3.0 (video generation, final frame)} & off & 0 (any frame: 1.7) & 0 & 0 & 0 & 0 & 0 & 0 \\
 & on & 83.3 (any frame: 91.7) & 90 & 80 & 90 & 80 & 70 & 90 \\
\bottomrule
\end{tabular}}
\caption{One switch with everything else fixed. The same canvas, the same fixed prompt, the same seed, and the same model are used, with the model's own prompt rewriting, prompt\_extend, turned off or on. Accuracy is exact and in percent. Wan 3.0 starts from the canvas as its first frame, generates 5 s of video, and is judged on 8 sampled frames. With the rewriter off, the two image editing models leave the answer box empty on every item, and Wan 3.0 leaves it empty or writes scribbles such as ``17'' or ``\#ab)''.}
\label{tab:t4}
\end{table}

\section{Post-training the generation module}
\label{app:C}

For FLUX.2-klein, FLUX.3-klein, Qwen-Image-2.1, and LongCat-Image-Edit, we train one LoRA per task on the attention and MLP projections of the diffusion transformer, with rank 16 and 2,500 updates, giving the answer box a larger weight in the flow-matching loss. The target is the same canvas with the answer written in. We also train a rendering LoRA the same way on items whose answer is written in the prompt, teaching the model to write a given answer rather than solve the question. With this rendering LoRA and the answer in the prompt, FLUX.2-klein, FLUX.3-klein, Qwen-Image-2.1, and LongCat-Image-Edit write the correct answer on 67.3\%, 99.0\%, 93.0\%, and 52.5\% of items across six test sets. The generators therefore can write, and what post-training fails to add is reasoning. On \emph{Instruction}, FLUX.3-klein writes a new word backwards, one of the five trained rules, on 92\% of items, whereas it drops the first letter, an unseen rule, on 0\%. The pattern in \Cref{tab:t5} holds even with more data, a higher rank, a LoRA placed on the connector or on the text encoder, or a curriculum that trains copying first. \textbf{BAGEL.} Its text module answers 82.9\% of 450 items over fifteen test sets in text tokens, reaching 89\% on Factual and 97\% on Arithmetic. Training either the generation expert alone or both experts together to write the answer on the canvas leaves Default accuracy between 0\% and 1.8\% on the same items, whereas the same training lifts Render accuracy from 46.7\% to 59.6\% when the answer is supplied in the prompt.

\begin{table}[!htb]
\centering
\small
\resizebox{\ifdim\width>\linewidth\linewidth\else\width\fi}{!}{%
\begin{tabular}{lcccc}
\toprule
\textbf{Task, set} & \textbf{FLUX.2-klein} & \textbf{FLUX.3-klein} & \textbf{Qwen-Image-2.1} & \textbf{LongCat-Image-Edit} \\
\midrule
Copy, new strings & 79.5 & 92.5 & 91.5 & 81.0 \\
Instruction, new words / new rule & 10.0 / 0 & 81.7 / 0 & 11.7 / 0 & 10.0 / 0 \\
Instruction, training items & 0 & 83.0 & 0 & 0 \\
Context, new names & 8.0 & 28.5 & 9.0 & 0.5 \\
Context, training items & 8.0 & 32.0 & 5.0 & 0 \\
Visual, new scenes & 6.5 & 7.0 & 3.5 & 8.0 \\
Visual, training items & 7.0 & 9.0 & 2.0 & 6.0 \\
Arithmetic, new expressions & 0 & 0.5 & 0.5 & 1.5 \\
Arithmetic, training items & 0 & 2.0 & 0 & 2.0 \\
Factual, paraphrased, reversed, two-hop, and unseen & 0 & 0 & 0 & 0 \\
Factual, training items & 0 & 0 & 0 & 0 \\
\bottomrule
\end{tabular}}
\caption{The numbers of \Cref{fig:f3}, for post-training of the generation module, as exact accuracy in percent on the test sets with an MLLM judge. Every base model scores 0 on every row. The Instruction row gives new words and a rule never trained. The training-item rows test recall on 100 trained items per task, each seen about 19 times, and match the test rows, confirming that post-training teaches copying of visible text rather than reasoning.}
\label{tab:t5}
\end{table}

\section{Where the answer is in the text module}
\label{app:D}

\textbf{Text modules as chat models.} We ask the text encoders of the four image editing models the questions of the benchmark and score their text-token answers. The encoders of FLUX.3-klein, Qwen-Image-2.1, and LongCat-Image-Edit read the canvas, whereas the encoder of FLUX.2-klein, which takes no image, reads the question as text. They answer 86\% to 88\% of paraphrased Factual questions and 83\% to 100\% of Arithmetic items, while the generators conditioned on them answer none. \textbf{Decoding the conditioning states.} In Qwen-Image-2.1, we apply the encoder's own language-model head to the hidden states that it passes to the diffusion transformer~\citep{nostalgebraist2020logitlens}. For 140 of the 150 evaluated Factual questions, the first token of the correct answer ranks among the 10 most probable tokens, yet the generator writes the answer for none of them. \textbf{Thoughts of unified models.} In Think mode we inspect each thought to check whether it contains the answer, then compare with the judged image. For SenseNova, the thought contains the answer on 290 of the 300 evaluated items, and the image is correct on 287 of those 290 items, or 99.0\%, compared with 1 of the remaining 10. For InternVL-U, the thought contains the answer on 148 items, and the image is correct on 36 of those, compared with none of the remaining 152. The image thus follows the thought, and InternVL-U writes even a supplied answer less reliably, as its Render accuracy of 1.3\% in \Cref{tab:t1} shows. \textbf{Emu3.5.} When queried in text tokens, 34.3\% of its responses contain the correct answer. In its thinking mode, the image generation module does not see the text-token answer, and the image is correct on only 0.5\% of items, confirming that without the answer in text tokens its generation module does not solve the questions itself.

\section{The tasks of the pixel models}
\label{app:E}

Training problems are generated on the fly during training, while test sets are fixed and constructed to prevent memorization. \emph{Expressions} have operands from 1 to 9 with probability 80\% and from 10 to 20 otherwise, with 1 to 4 operations in training and 1, 2, 3, 4, 6, and 8 operations in testing. Operators are drawn from +, \ensuremath{-}, and \ensuremath{\times}, and an expression is resampled if any intermediate value is negative or it does not fit on one row. Each frame performs one operation, the leftmost multiplication first and then from left to right. \emph{Multiplication} uses up to 3\ensuremath{\times}3 digits in training, exactly 3\ensuremath{\times}3 digits in testing, and 4\ensuremath{\times}4 digits in the scaling ladder. Each frame writes one partial product, right aligned and shifted, and the last frame writes the sum. In \emph{Word Problems}, a person starts with 2 to 60 objects, and each sentence adds, removes, multiplies, or halves them. Training uses made-up names, whereas testing uses real names never seen during training. \emph{Word Problems} use a 256\ensuremath{\times}256 canvas with 32 columns and 16 rows of cells and a batch size of 128, since three-step problems require more rows. \emph{Factual Questions} use a knowledge base of 300 people, each assigned a city, a job, and a mother, with cities belonging to countries. All single facts are included in training. Two-hop questions about the country, the mother's city, or the mother's job, along with three-hop questions about the mother's country, are trained on half of the people and tested on the remaining half. In \emph{Curve Tracing}, K curves connect letters on the left to numbers on the right and cross each other, with each frame inking a quarter of the queried curve. In \emph{Trace Then Sum}, the question, such as ``B+E=?'', asks for the sum of the numbers reached by the curves starting at the two letters, and after inking each curve the model records notes such as ``B=4'' and ``E=7'' on the note row just above the horizontal rule before the last frame writes the sum. \emph{Mazes} are n\ensuremath{\times}n grids of free cells and holes, with $3 \le n \le 5$, a start S, and a goal G, under the question ``Path from S to G:''. Each frame inks one more cell of the path and writes its move, R, D, L, or U, on the note row, and the last frame writes all moves in the answer box. The default demonstration follows the breadth-first shortest path with ties broken in a fixed order, whereas direct answering draws the entire path and writes all moves in one frame. \emph{Sum Mazes} place a digit from 1 to 9 in every free cell except S and G. Each frame advances one cell along a shortest path, writes the move on the note row, and updates the running total, and the last frame writes the total in the answer box. \emph{Count Then Paint} shows three irregular regions in distinct greys, each labeled with a letter and containing 1 to 6 dots. Each frame circles one dot and updates the count, as in A1, A2, A3, A3 B1, and so on, and the final frames write the chosen letter, box the label with that letter, and paint the boxed region. During training, counterfactual letters randomly replace the written letter and set the target to paint the corresponding region, tying the painted region directly to the written letter. In \emph{Compute Then Paint}, the question is an expression such as ``paint where dots = 7\ensuremath{-}2''. The first frame writes the result 7\ensuremath{-}2=5, and each following frame circles one dot and updates the count, as in A1, A2, \ldots{}, A5 B1 C3. Once counting finishes, one frame per region appends + if its count equals the target result and \ensuremath{-} otherwise, as in A\ensuremath{-} B+ C\ensuremath{-}, and the answer box copies the letter of the region marked +. The remaining frames box that label and paint its region, so a problem takes 13 to 22 frames, or 17.7 on average.

\section{Training}
\label{app:F}

All pixel models use the recipe of \Cref{tab:t6}. The 458M models are trained with data parallelism across 2 to 4 GPUs while keeping the same global batch size. In frame-by-frame training, the input canvas is perturbed with Gaussian noise of standard deviation 0.03 so that rollouts tolerate their own small errors. The scaling ladders train for 132k steps, with the learning rate decayed linearly to zero over the last 4k steps. In \Cref{sec:6}, models compared under an equal training budget run for 40k steps. After warm-up, the learning rate stays constant on \emph{Expressions} and \emph{Word Problems}, cools down starting at 20k steps on \emph{Count Then Paint}, and cools down starting at 36k steps on \emph{Sum Mazes}, \emph{Trace Then Sum}, and \emph{Compute Then Paint}. The longer runs train for 160k steps with a cool-down from 152k steps. \emph{Trace Then Sum} and \emph{Compute Then Paint} initialize from the \emph{Curve Tracing} and \emph{Count Then Paint} models of the same answering mode, the former trained for 40k steps at a constant learning rate. On the cross-modal tasks, the loss weights pixels changed by a step 31 times higher than copied pixels, except in the \emph{Curve Tracing} model that initializes \emph{Trace Then Sum}. The joint cross-modal ladder draws 88 examples from each of \emph{Mazes}, \emph{Curve Tracing}, and \emph{Count Then Paint} in every step, forming a batch of 264.

\begin{table}[!htb]
\centering
\small
\resizebox{\ifdim\width>\linewidth\linewidth\else\width\fi}{!}{%
\begin{tabular}{lccccccc}
\toprule
\textbf{Size} & \textbf{XS} & \textbf{S} & \textbf{M} & \textbf{B} & \textbf{L} \\
\midrule
Width / depth / heads & 256 / 6 / 4 & 384 / 12 / 6 & 512 / 12 / 8 & 768 / 12 / 12 & 1024 / 24 / 16 \\
Parameters & 7.6M & 33M & 58M & 130M & 458M \\
\bottomrule
\end{tabular}}
\caption{Model sizes. All sizes use the same recipe, with batch 256, 128 on \emph{Word Problems}, AdamW with \ensuremath{\beta} of 0.9 and 0.95, learning rate 10\textsuperscript{\ensuremath{-}4} with 1,000 warm-up steps, EMA 0.999, and bf16, for 40k steps in Sections 6 and 7 and 132k steps in the scaling ladders of \Cref{sec:5}.}
\label{tab:t6}
\end{table}

\section{Evaluation of the pixel models}
\label{app:G}

Answers are read directly from pixels, with no judge model. Each 8\ensuremath{\times}16 cell of the answer rows is matched to the nearest glyph of the font, and an answer is correct only if the whole string matches. Frame-by-frame rollouts start from the original problem canvas and generate each frame from the previous one using the deterministic Euler sampler, with 30 steps on the text tasks and 10 steps on the cross-modal tasks as well as for the per-step error of \Cref{sec:5}. In \Cref{sec:6}, each frame-by-frame rollout runs for the frame count of the demonstration and the answer is read from the last frame, except on \emph{Sum Mazes}, where the rollout stops at the first frame whose answer rows are not blank, or after 22 frames. In \Cref{sec:7}, where the model chooses the number of frames, every rollout stops at the first frame with a non-blank answer, or upon reaching a cap of 10 frames on \emph{Expressions}, 8 on \emph{Word Problems} for GRPO and 10 for the other runs, 22 on \emph{Sum Mazes}, and 32 on the other cross-modal tasks. Direct answering generates a single frame. \emph{Trace Then Sum} is marked correct when the sum in the answer box is exact, and \emph{Sum Mazes} is correct when the moves written on the note row match the demonstrated path and the answer box gives its exact total. \emph{Count Then Paint} and \emph{Compute Then Paint} are correct when the letter is right, the note row is right, which for \emph{Compute Then Paint} includes the arithmetic result and the + and \ensuremath{-} marks, and the painted pixels overlap the true region with an IoU of at least 0.8. Direct answering writes no counts, so only the letter and the painting are checked. In the RL comparison of \Cref{sec:7}, a cross-modal problem counts as solved only when it is solved in the fewest frames, which on \emph{Sum Mazes} requires finding a shortest path and writing its own total in the answer box. Every accuracy of the pixel models in the main text is measured on 1,024 problems per test set generated with a fixed seed, so all compared runs see the same problems. The per-step error of \Cref{sec:5} uses 4,096 fixed transitions per test set, where the model generates the next frame from a correct canvas. On text tasks a step is counted as wrong if any cell is read back incorrectly, and on \emph{Curve Tracing} it is wrong if the changed pixels have an IoU below 0.8 with the true change.

\section{Reinforcement learning}
\label{app:H}

\textbf{Sampler.} With $t = 0$ for noise and $t = 1$ for data, $x_t = t\,y + (1-t)\,\epsilon$, and the model predicts y. The rollout sampler is the SDE $dx = [v - \tfrac{g^2}{2}\,\hat\epsilon/(1-t)]\,dt + g\,dW$ with $v = (\hat y - x)/(1-t)$, $\hat\epsilon = (x - t\,\hat y)/(1-t)$, and $g = 0.7\,(1-t)$, which shares the same marginals as the deterministic sampler~\citep{liu2025flowgrpo}, following Flow-GRPO and earlier policy-gradient training of diffusion models~\citep{black2024ddpo}, so every frame has tractable transition probabilities. Each frame uses 10 SDE steps, and the last step returns $\hat y$ without noise. \textbf{GRPO.} Each iteration samples 8 rollouts for each of 16 problems, with at most 10 frames each, or 8 on \emph{Word Problems}. The reward is 1 for an exact answer plus 0.2 times the fraction of correct characters. Advantages are normalized within each group and clipped at \ensuremath{\pm}5, and every transition of a rollout shares its advantage. All SDE transitions of an iteration, at most 2,048, are accumulated into one gradient step with learning rate 10\textsuperscript{\ensuremath{-}5} and EMA 0.99, with a KL penalty of weight 0.004 to the supervised model, for 1,500 iterations. Training problems come from the hardest level of each task, excluding the test problems. \textbf{Controls.} The same compute in \Cref{sec:7} means that non-optimal and optimal SFT match RL in iteration count or training steps. On the cross-modal tasks, all three methods run for 600 iterations, where SFT takes one update on 128 frame pairs per iteration and RFT takes up to four updates on at most 512 redrawn pairs in batches of 128. On the text tasks, SFT continues for 25k steps, while GRPO runs 1,500 iterations of one update each. The demonstrations are described below. \textbf{RFT.} On \emph{Sum Mazes}, \emph{Trace Then Sum}, \emph{Count Then Paint}, and \emph{Compute Then Paint}, each iteration keeps the correct rollouts with the fewest frames. On \emph{Sum Mazes} these reach the goal by a shortest path, and on \emph{Trace Then Sum} they also write both intermediate values right. Across 600 iterations, each iteration redraws those selected rollouts as clean frames matching what the model wrote and trains on them with the supervised loss, using learning rate 10\textsuperscript{\ensuremath{-}5}, EMA 0.99, and 8 samples for each of 16 problems per iteration. Evaluation uses the EMA weights of each RL run.

\textbf{Demonstrations.} On the cross-modal tasks, non-optimal demonstrations take detours, and the start models are trained on them for 40k steps with changed pixels weighted as in Appendix~\ref{app:F}. In \emph{Sum Mazes} they follow a longer path, in \emph{Count Then Paint} and \emph{Compute Then Paint} half of them first box a wrong region and correct it in the next frame, and in \emph{Trace Then Sum} half of them first trace a curve the question does not ask about. Optimal demonstrations take the fewest frames and no detour. On the text tasks, non-optimal demonstrations repeat each step on a new row with probability one half, whereas optimal demonstrations carry out one to three randomly chosen operations in each frame.

\section{Further results}
\label{app:I}

\textbf{Word problems across sizes.} On three-step \emph{Word Problems} with new names, frame by frame solves 3.7\%, 47.8\%, and 78.6\% at 7.6M, 33M, and 130M after 40k steps, while direct answering solves 0.5\%, 1.1\%, and 1.2\%, each evaluated on 1,024 problems. Thus, the ability to solve them emerges with scale only when the steps are written frame by frame, corresponding to the emergence that \Cref{sec:6} refers to.

\section{Scaling fits}
\label{app:J}

\begin{table}[!htb]
\centering
\small
\resizebox{\ifdim\width>\linewidth\linewidth\else\width\fi}{!}{%
\begin{tabular}{lllllll}
\toprule
\textbf{Curve} & \textbf{Fit on 7.6M to 130M} & \textbf{Log RMSE} & \textbf{\makecell[l]{458M predicted\\(90\% interval)}} & \textbf{\makecell[l]{458M measured\\(90\% interval)}} & \textbf{\makecell[l]{Relative error\\(90\% interval)}} & \textbf{Fit on all five sizes} \\
\midrule
Text tasks & 49.6\ensuremath{\cdot}N\textsuperscript{\ensuremath{-}0.360} & 0.056 & \makecell[l]{0.0381\\(0.0337--0.0443)} & \makecell[l]{0.0374\\(0.0271--0.0442)} & \makecell[l]{\ensuremath{-}2\%\\(\ensuremath{-}26\% to +23\%)} & \makecell[l]{52.6\ensuremath{\cdot}N\textsuperscript{\ensuremath{-}0.363}\\(0.050)} \\
\emph{Curve Tracing}, joint model & 111\ensuremath{\cdot}N\textsuperscript{\ensuremath{-}0.340} & 0.052 & \makecell[l]{0.1259\\(0.1126--0.1335)} & \makecell[l]{0.1359\\(0.1234--0.1559)} & \makecell[l]{+8\%\\(\ensuremath{-}4\% to +24\%)} & \makecell[l]{84.0\ensuremath{\cdot}N\textsuperscript{\ensuremath{-}0.324}\\(0.051)} \\
\bottomrule
\end{tabular}}
\caption{Power-law fits of the per-step error of \Cref{sec:5}, on the medians of five seeds per size. The last column fits all five sizes and gives the log RMSE after the fit.}
\label{tab:t7}
\end{table}

\textbf{Method.} The per-step error of the text tasks is the mean over four-step \emph{Expressions}, 4\ensuremath{\times}4-digit \emph{Multiplication}, and \emph{Factual Questions}, the last averaged over four sets of new entities. Each test set uses 4,096 transitions. On \emph{Curve Tracing}, it is the share of transitions whose changed pixels have an IoU below 0.8 with the true change, using 4,096 transitions per test set averaged over the sets with 2, 4, 6, and 8 curves. For each size we take the median over five seeds and fit a straight line to log error against log non-embedding parameters N, so that the error follows A\ensuremath{\cdot}N\textsuperscript{\ensuremath{-}\ensuremath{\alpha}}. The held-out check fits the power law only on sizes from 7.6M to 130M and extrapolates to 458M, with relative error defined as measured / predicted \ensuremath{-} 1. The 90\% interval comes from a seed bootstrap with 4,000 draws. In each draw we resample the five seeds of every size with replacement, recompute the medians, fit again, and predict 458M, and the resulting interval spans the 5th to the 95th percentile of the predictions. In the same way, we resample the five seeds of the 458M model to obtain the 90\% interval of its measured median, and resample all sizes jointly to obtain the 90\% interval of the relative error. Because the measured 458M median lies inside the predicted 90\% interval for the text tasks and the two intervals overlap for \emph{Curve Tracing}, the law fitted on models up to 130M accurately predicts the held-out 458M model.

%% file: main.bbl
\begin{thebibliography}{48}
\providecommand{\natexlab}[1]{#1}
\providecommand{\url}[1]{\texttt{#1}}
\expandafter\ifx\csname urlstyle\endcsname\relax
  \providecommand{\doi}[1]{doi: #1}\else
  \providecommand{\doi}{doi: \begingroup \urlstyle{rm}\Url}\fi

\bibitem[Betker et~al.(2023)Betker, Goh, Jing, Brooks, Wang, Li, Ouyang,
  Zhuang, Lee, Guo, Manassra, Dhariwal, Chu, Jiao, and
  Ramesh]{betker2023dalle3}
James Betker, Gabriel Goh, Li~Jing, Tim Brooks, Jianfeng Wang, Linjie Li, Long
  Ouyang, Juntang Zhuang, Joyce Lee, Yufei Guo, Wesam Manassra, Prafulla
  Dhariwal, Casey Chu, Yunxin Jiao, and Aditya Ramesh.
\newblock Improving image generation with better captions.
\newblock Technical report, OpenAI, 2023.

\bibitem[Black et~al.(2024)Black, Janner, Du, Kostrikov, and
  Levine]{black2024ddpo}
Kevin Black, Michael Janner, Yilun Du, Ilya Kostrikov, and Sergey Levine.
\newblock Training diffusion models with reinforcement learning.
\newblock In \emph{The Twelfth International Conference on Learning
  Representations}, 2024.

\bibitem[Brooks et~al.(2023)Brooks, Holynski, and
  Efros]{brooks2023instructpix2pix}
Tim Brooks, Aleksander Holynski, and Alexei~A. Efros.
\newblock Instructpix2pix: Learning to follow image editing instructions, 2023.

\bibitem[Chai et~al.(2024)Chai, Liu, Xiao, Wang, Sun, and Wu]{chai2024pixelgpt}
Yekun Chai, Qingyi Liu, Jingwu Xiao, Shuohuan Wang, Yu~Sun, and Hua Wu.
\newblock Autoregressive pre-training on pixels and texts.
\newblock In \emph{Proceedings of the 2024 Conference on Empirical Methods in
  Natural Language Processing}, pages 3106--3125, Miami, Florida, USA, 2024.
  Association for Computational Linguistics.

\bibitem[Chen et~al.(2023)Chen, Huang, Lv, Cui, Chen, and
  Wei]{chen2023textdiffuser}
Jingye Chen, Yupan Huang, Tengchao Lv, Lei Cui, Qifeng Chen, and Furu Wei.
\newblock Textdiffuser: Diffusion models as text painters.
\newblock In \emph{Thirty-seventh Conference on Neural Information Processing
  Systems}, 2023.

\bibitem[Chen et~al.(2025)Chen, Wu, Liu, Pan, Liu, Xie, Yu, and
  Ruan]{chen2025janus}
Xiaokang Chen, Zhiyu Wu, Xingchao Liu, Zizheng Pan, Wen Liu, Zhenda Xie,
  Xingkai Yu, and Chong Ruan.
\newblock Janus-pro: Unified multimodal understanding and generation with data
  and model scaling, 2025.

\bibitem[Deng et~al.(2025)Deng, Zhu, Li, Gou, Li, Wang, Zhong, Yu, Nie, Song,
  Shi, and Fan]{deng2025bagel}
Chaorui Deng, Deyao Zhu, Kunchang Li, Chenhui Gou, Feng Li, Zeyu Wang, Shu
  Zhong, Weihao Yu, Xiaonan Nie, Ziang Song, Guang Shi, and Haoqi Fan.
\newblock Emerging properties in unified multimodal pretraining, 2025.

\bibitem[Fang et~al.(2025)Fang, Duan, Wang, Huang, Li, Tian, Yan, Yu, Zeng,
  Dai, Liu, and Li]{fang2025got}
Rongyao Fang, Chengqi Duan, Kun Wang, Linjiang Huang, Hao Li, Hao Tian, Shilin
  Yan, Weihao Yu, Xingyu Zeng, Jifeng Dai, Xihui Liu, and Hongsheng Li.
\newblock {GoT}: Unleashing reasoning capability of {MLLM} for visual
  generation and editing.
\newblock In \emph{The Thirty-ninth Annual Conference on Neural Information
  Processing Systems}, 2025.

\bibitem[{Google}()]{google2026geminithinking}
{Google}.
\newblock Gemini thinking.
\newblock Gemini {API} documentation.
\newblock Accessed: 2026-11-16.

\bibitem[Guo et~al.(2025)]{guo2025deepseekr1}
Daya Guo et~al.
\newblock {DeepSeek-R1}: Incentivizing reasoning capability in {LLMs} via
  reinforcement learning.
\newblock \emph{arXiv preprint arXiv:2501.12948}, 2025.

\bibitem[Guo et~al.(2026)Guo, Chen, Zhang, An, Qi, Jiang, Li, Zhang, Li, and
  Heng]{guo2025mmecof}
Ziyu Guo, Xinyan Chen, Renrui Zhang, Ruichuan An, Yu~Qi, Dongzhi Jiang,
  Xiangtai Li, Manyuan Zhang, Hongsheng Li, and Pheng-Ann Heng.
\newblock Are video models ready as zero-shot reasoners? an empirical study
  with the {MME-CoF} benchmark.
\newblock In \emph{Conference on Computer Vision and Pattern Recognition 2026},
  2026.

\bibitem[Hoffmann et~al.(2022)Hoffmann, Borgeaud, Mensch, Buchatskaya, Cai,
  Rutherford, de~Las~Casas, Hendricks, Welbl, Clark, Hennigan, Noland,
  Millican, van~den Driessche, Damoc, Guy, Osindero, Simonyan, Elsen, Vinyals,
  Rae, and Sifre]{hoffmann2022chinchilla}
Jordan Hoffmann, Sebastian Borgeaud, Arthur Mensch, Elena Buchatskaya, Trevor
  Cai, Eliza Rutherford, Diego de~Las~Casas, Lisa~Anne Hendricks, Johannes
  Welbl, Aidan Clark, Tom Hennigan, Eric Noland, Katie Millican, George van~den
  Driessche, Bogdan Damoc, Aurelia Guy, Simon Osindero, Karen Simonyan, Erich
  Elsen, Oriol Vinyals, Jack~W. Rae, and Laurent Sifre.
\newblock Training compute-optimal large language models.
\newblock In \emph{Advances in Neural Information Processing Systems}, pages
  30016--30030. Curran Associates, Inc., 2022.

\bibitem[Hu et~al.(2024)Hu, Shi, Fu, Roth, Ostendorf, Zettlemoyer, Smith, and
  Krishna]{hu2024sketchpad}
Yushi Hu, Weijia Shi, Xingyu Fu, Dan Roth, Mari Ostendorf, Luke Zettlemoyer,
  Noah~A. Smith, and Ranjay Krishna.
\newblock Visual sketchpad: Sketching as a visual chain of thought for
  multimodal language models.
\newblock In \emph{The Thirty-eighth Annual Conference on Neural Information
  Processing Systems}, 2024.

\bibitem[Kaplan et~al.(2020)Kaplan, McCandlish, Henighan, Brown, Chess, Child,
  Gray, Radford, Wu, and Amodei]{kaplan2020scaling}
Jared Kaplan, Sam McCandlish, Tom Henighan, Tom~B. Brown, Benjamin Chess, Rewon
  Child, Scott Gray, Alec Radford, Jeffrey Wu, and Dario Amodei.
\newblock Scaling laws for neural language models, 2020.

\bibitem[Kou et~al.(2026)Kou, Jin, Zhou, Ma, Wang, Chen, Jiang, Yang, Zhu, Yu,
  and Deng]{kou2026think}
Siqi Kou, Jiachun Jin, Zetong Zhou, Ye~Ma, Yugang Wang, Quan Chen, Peng Jiang,
  Xiao Yang, Jun Zhu, Kai Yu, and Zhijie Deng.
\newblock Think-then-generate: Reasoning-aware text-to-image diffusion with
  {LLM} encoders.
\newblock In \emph{Forty-third International Conference on Machine Learning},
  2026.

\bibitem[Labs et~al.(2025)Labs, Batifol, Blattmann, Boesel, Consul, Diagne,
  Dockhorn, English, English, Esser, Kulal, Lacey, Levi, Li, Lorenz,
  M{\"u}ller, Podell, Rombach, Saini, Sauer, and Smith]{labs2025kontext}
Black~Forest Labs, Stephen Batifol, Andreas Blattmann, Frederic Boesel, Saksham
  Consul, Cyril Diagne, Tim Dockhorn, Jack English, Zion English, Patrick
  Esser, Sumith Kulal, Kyle Lacey, Yam Levi, Cheng Li, Dominik Lorenz, Jonas
  M{\"u}ller, Dustin Podell, Robin Rombach, Harry Saini, Axel Sauer, and Luke
  Smith.
\newblock {FLUX.1 Kontext}: Flow matching for in-context image generation and
  editing in latent space, 2025.

\bibitem[Li et~al.(2025)Li, Wu, Zhang, Xia, Mao, Dong, Vuli{\'c}, and
  Wei]{li2025mvot}
Chengzu Li, Wenshan Wu, Huanyu Zhang, Yan Xia, Shaoguang Mao, Li~Dong, Ivan
  Vuli{\'c}, and Furu Wei.
\newblock Imagine while reasoning in space: Multimodal
  visualization-of-thought.
\newblock In \emph{Forty-second International Conference on Machine Learning},
  2025.

\bibitem[Li and He(2026)]{li2025jit}
Tianhong Li and Kaiming He.
\newblock Back to basics: Let denoising generative models denoise.
\newblock In \emph{Proceedings of the IEEE/CVF Conference on Computer Vision
  and Pattern Recognition (CVPR)}, 2026.

\bibitem[Liang et~al.(2026)Liang, He, Yang, and Dai]{liang2024scalingdit}
Zhengyang Liang, Hao He, Ceyuan Yang, and Bo~Dai.
\newblock Scaling laws for diffusion transformers, 2026.

\bibitem[Lipman et~al.(2023)Lipman, Chen, Ben-Hamu, Nickel, and
  Le]{lipman2023flow}
Yaron Lipman, Ricky T.~Q. Chen, Heli Ben-Hamu, Maximilian Nickel, and Matthew
  Le.
\newblock Flow matching for generative modeling.
\newblock In \emph{The Eleventh International Conference on Learning
  Representations}, 2023.

\bibitem[Liu et~al.(2025)Liu, Liu, Liang, Li, Liu, Wang, Wan, Zhang, and
  Ouyang]{liu2025flowgrpo}
Jie Liu, Gongye Liu, Jiajun Liang, Yangguang Li, Jiaheng Liu, Xintao Wang,
  Pengfei Wan, Di~Zhang, and Wanli Ouyang.
\newblock {Flow-GRPO}: Training flow matching models via online {RL}.
\newblock In \emph{The Thirty-ninth Annual Conference on Neural Information
  Processing Systems}, 2025.

\bibitem[Liu et~al.(2023)Liu, Gong, and Liu]{liu2023rectified}
Xingchao Liu, Chengyue Gong, and Qiang Liu.
\newblock Flow straight and fast: Learning to generate and transfer data with
  rectified flow.
\newblock In \emph{The Eleventh International Conference on Learning
  Representations}, 2023.

\bibitem[Niu et~al.(2026)Niu, Ning, Zheng, Jin, Lin, Jin, Liao, Feng, Meng,
  Ning, Zhu, and Yuan]{niu2025wise}
Yuwei Niu, Munan Ning, Mengren Zheng, Weiyang Jin, Bin Lin, Peng Jin, Jiaqi
  Liao, Chaoran Feng, Fanqing Meng, Kun-Peng Ning, Bin Zhu, and Li~Yuan.
\newblock {WISE}: World knowledge-informed semantic evaluation for
  text-to-image generation.
\newblock In \emph{Forty-third International Conference on Machine Learning},
  2026.

\bibitem[nostalgebraist(2020)]{nostalgebraist2020logitlens}
nostalgebraist.
\newblock Interpreting {GPT}: the logit lens.
\newblock LessWrong blog post, 2020.

\bibitem[Nye et~al.(2021)Nye, Andreassen, Gur-Ari, Michalewski, Austin, Bieber,
  Dohan, Lewkowycz, Bosma, Luan, Sutton, and Odena]{nye2021scratchpad}
Maxwell Nye, Anders~Johan Andreassen, Guy Gur-Ari, Henryk Michalewski, Jacob
  Austin, David Bieber, David Dohan, Aitor Lewkowycz, Maarten Bosma, David
  Luan, Charles Sutton, and Augustus Odena.
\newblock Show your work: Scratchpads for intermediate computation with
  language models.
\newblock \emph{arXiv preprint arXiv:2112.00114}, 2021.

\bibitem[{OpenAI}()]{openai2026imagegen}
{OpenAI}.
\newblock Image generation.
\newblock {OpenAI} {API} documentation.
\newblock Accessed: 2026-11-16.

\bibitem[Ouyang et~al.(2022)]{ouyang2022instructgpt}
Long Ouyang et~al.
\newblock Training language models to follow instructions with human feedback.
\newblock In \emph{Advances in Neural Information Processing Systems
  (NeurIPS)}, pages 27730--27744, 2022.

\bibitem[Peebles and Xie(2023)]{peebles2023dit}
William Peebles and Saining Xie.
\newblock Scalable diffusion models with transformers.
\newblock In \emph{Proceedings of the IEEE/CVF International Conference on
  Computer Vision (ICCV)}, pages 4195--4205, 2023.

\bibitem[Ren et~al.(2026)Ren, Chen, Wang, Song, Zhu, Yuille, Chen, and
  Lu]{ren2026unireasoner}
Sucheng Ren, Chen Chen, Zhenbang Wang, Liangchen Song, Xiangxin Zhu, Alan
  Yuille, Liang-Chieh Chen, and Jiasen Lu.
\newblock Large language models are universal reasoners for visual generation,
  2026.

\bibitem[Rust et~al.(2023)Rust, Lotz, Bugliarello, Salesky, de~Lhoneux, and
  Elliott]{rust2023pixel}
Phillip Rust, Jonas~F. Lotz, Emanuele Bugliarello, Elizabeth Salesky, Miryam
  de~Lhoneux, and Desmond Elliott.
\newblock Language modelling with pixels.
\newblock In \emph{The Eleventh International Conference on Learning
  Representations}, 2023.

\bibitem[Shao et~al.(2024)Shao, Wang, Zhu, Xu, Song, Bi, Zhang, Zhang, Li, Wu,
  and Guo]{shao2024deepseekmath}
Zhihong Shao, Peiyi Wang, Qihao Zhu, Runxin Xu, Junxiao Song, Xiao Bi, Haowei
  Zhang, Mingchuan Zhang, Y.~K. Li, Y.~Wu, and Daya Guo.
\newblock {DeepSeekMath}: Pushing the limits of mathematical reasoning in open
  language models, 2024.

\bibitem[Sun et~al.(2026)Sun, Fang, Duan, Liu, Li, and
  Liu]{sun2025t2ireasonbench}
Kaiyue Sun, Rongyao Fang, Chengqi Duan, Xian Liu, Aoxue Li, and Xihui Liu.
\newblock {T2I-ReasonBench}: Benchmarking reasoning-informed text-to-image
  generation, 2026.

\bibitem[Tai et~al.(2024)Tai, Liao, Suglia, and Vergari]{tai2024pixar}
Yintao Tai, Xiyang Liao, Alessandro Suglia, and Antonio Vergari.
\newblock {PIXAR}: Auto-regressive language modeling in pixel space.
\newblock In \emph{Findings of the Association for Computational Linguistics:
  ACL 2024}, pages 14673--14695, Bangkok, Thailand, 2024. Association for
  Computational Linguistics.

\bibitem[Team(2025)]{team2024chameleon}
Chameleon Team.
\newblock Chameleon: Mixed-modal early-fusion foundation models, 2025.

\bibitem[Tong et~al.(2026)Tong, Mou, Li, Li, Yang, Zhang, Chen, Liang, Hu,
  Zheng, Chen, Zhao, Huang, and Qiu]{tong2025thinkingvideo}
Jingqi Tong, Yurong Mou, Hangcheng Li, Mingzhe Li, Yongzhuo Yang, Ming Zhang,
  Qiguang Chen, Tianyi Liang, Xiaomeng Hu, Yining Zheng, Xinchi Chen, Jun Zhao,
  Xuanjing Huang, and Xipeng Qiu.
\newblock Thinking with video: Video generation as a promising multimodal
  reasoning paradigm.
\newblock In \emph{Conference on Computer Vision and Pattern Recognition 2026},
  2026.

\bibitem[Tschannen et~al.(2023)Tschannen, Mustafa, and
  Houlsby]{tschannen2023clippo}
Michael Tschannen, Basil Mustafa, and Neil Houlsby.
\newblock {CLIPPO}: Image-and-language understanding from pixels only.
\newblock In \emph{2023 IEEE/CVF Conference on Computer Vision and Pattern
  Recognition (CVPR)}, pages 11006--11017, 2023.

\bibitem[Wan et~al.(2025)Wan, Wang, Ai, Wen, Mao, Xie, Chen, Yu, Zhao, Yang,
  Zeng, Wang, Zhang, Zhou, Wang, Chen, Zhu, Zhao, Yan, Huang, Feng, Zhang, Li,
  Wu, Chu, Feng, Zhang, Sun, Fang, Wang, Gui, Weng, Shen, Lin, Wang, Wang,
  Zhou, Wang, Shen, Yu, Shi, Huang, Xu, Kou, Lv, Li, Liu, Wang, Zhang, Huang,
  Li, Wu, Liu, Pan, Zheng, Hong, Shi, Feng, Jiang, Han, Wu, and
  Liu]{wan2025wan}
Team Wan, Ang Wang, Baole Ai, Bin Wen, Chaojie Mao, Chen-Wei Xie, Di~Chen,
  Feiwu Yu, Haiming Zhao, Jianxiao Yang, Jianyuan Zeng, Jiayu Wang, Jingfeng
  Zhang, Jingren Zhou, Jinkai Wang, Jixuan Chen, Kai Zhu, Kang Zhao, Keyu Yan,
  Lianghua Huang, Mengyang Feng, Ningyi Zhang, Pandeng Li, Pingyu Wu, Ruihang
  Chu, Ruili Feng, Shiwei Zhang, Siyang Sun, Tao Fang, Tianxing Wang, Tianyi
  Gui, Tingyu Weng, Tong Shen, Wei Lin, Wei Wang, Wei Wang, Wenmeng Zhou, Wente
  Wang, Wenting Shen, Wenyuan Yu, Xianzhong Shi, Xiaoming Huang, Xin Xu, Yan
  Kou, Yangyu Lv, Yifei Li, Yijing Liu, Yiming Wang, Yingya Zhang, Yitong
  Huang, Yong Li, You Wu, Yu~Liu, Yulin Pan, Yun Zheng, Yuntao Hong, Yupeng
  Shi, Yutong Feng, Zeyinzi Jiang, Zhen Han, Zhi-Fan Wu, and Ziyu Liu.
\newblock Wan: Open and advanced large-scale video generative models, 2025.

\bibitem[Wei et~al.(2025)Wei, Sun, and Li]{wei2025deepseekocr}
Haoran Wei, Yaofeng Sun, and Yukun Li.
\newblock {DeepSeek-OCR}: Contexts optical compression, 2025.

\bibitem[Wei et~al.(2022{\natexlab{a}})Wei, Tay, Bommasani, Raffel, Zoph,
  Borgeaud, Yogatama, Bosma, Zhou, Metzler, Chi, Hashimoto, Vinyals, Liang,
  Dean, and Fedus]{wei2022emergent}
Jason Wei, Yi~Tay, Rishi Bommasani, Colin Raffel, Barret Zoph, Sebastian
  Borgeaud, Dani Yogatama, Maarten Bosma, Denny Zhou, Donald Metzler, Ed~H.
  Chi, Tatsunori Hashimoto, Oriol Vinyals, Percy Liang, Jeff Dean, and William
  Fedus.
\newblock Emergent abilities of large language models.
\newblock \emph{Transactions on Machine Learning Research}, 2022{\natexlab{a}}.
\newblock Survey Certification.

\bibitem[Wei et~al.(2022{\natexlab{b}})Wei, Wang, Schuurmans, Bosma, Ichter,
  Xia, Chi, Le, and Zhou]{wei2022cot}
Jason Wei, Xuezhi Wang, Dale Schuurmans, Maarten Bosma, Brian Ichter, Fei Xia,
  Ed~H. Chi, Quoc~V Le, and Denny Zhou.
\newblock Chain-of-thought prompting elicits reasoning in large language
  models.
\newblock In \emph{Advances in Neural Information Processing Systems},
  2022{\natexlab{b}}.

\bibitem[Wiedemer et~al.(2025)Wiedemer, Li, Vicol, Gu, Matarese, Swersky, Kim,
  Jaini, and Geirhos]{wiedemer2025video}
Thadd{\"a}us Wiedemer, Yuxuan Li, Paul Vicol, Shixiang~Shane Gu, Nick Matarese,
  Kevin Swersky, Been Kim, Priyank Jaini, and Robert Geirhos.
\newblock Video models are zero-shot learners and reasoners, 2025.

\bibitem[Wu et~al.(2025)Wu, Li, Zhou, Lin, Gao, Yan, Yin, Bai, Xu, Chen, Chen,
  Tang, Zhang, Wang, Yang, Yu, Cheng, Liu, Li, Zhang, Meng, Wei, Ni, Chen, Cao,
  Peng, Qu, Wu, Wang, Yu, Wen, Feng, Xu, Wang, Zhang, Zhu, Wu, Cai, and
  Liu]{wu2025qwenimage}
Chenfei Wu, Jiahao Li, Jingren Zhou, Junyang Lin, Kaiyuan Gao, Kun Yan,
  Shengming Yin, Shuai Bai, Xiao Xu, Yilei Chen, Yuxiang Chen, Zecheng Tang,
  Zekai Zhang, Zhengyi Wang, An~Yang, Bowen Yu, Chen Cheng, Dayiheng Liu,
  Deqing Li, Hang Zhang, Hao Meng, Hu~Wei, Jingyuan Ni, Kai Chen, Kuan Cao,
  Liang Peng, Lin Qu, Minggang Wu, Peng Wang, Shuting Yu, Tingkun Wen, Wensen
  Feng, Xiaoxiao Xu, Yi~Wang, Yichang Zhang, Yongqiang Zhu, Yujia Wu, Yuxuan
  Cai, and Zenan Liu.
\newblock Qwen-image technical report, 2025.

\bibitem[Wu et~al.(2026)Wu, Li, Hu, Ye, Zeng, Yu, Zhu, Schiele, Yang, and
  Yang]{wu2025krisbench}
Yongliang Wu, Zonghui Li, Xinting Hu, Xinyu Ye, Xianfang Zeng, Gang Yu, Wenbo
  Zhu, Bernt Schiele, Ming-Hsuan Yang, and Xu~Yang.
\newblock {KRIS-Bench}: Benchmarking next-level intelligent image editing
  models.
\newblock In \emph{The Thirty-ninth Annual Conference on Neural Information
  Processing Systems Datasets and Benchmarks Track}, 2026.

\bibitem[Xie et~al.(2025)Xie, Mao, Bai, Zhang, Wang, Lin, Gu, Chen, Yang, and
  Shou]{xie2024showo}
Jinheng Xie, Weijia Mao, Zechen Bai, David~Junhao Zhang, Weihao Wang,
  Kevin~Qinghong Lin, Yuchao Gu, Zhijie Chen, Zhenheng Yang, and Mike~Zheng
  Shou.
\newblock Show-o: One single transformer to unify multimodal understanding and
  generation, 2025.

\bibitem[Yan et~al.(2025)Yan, Ye, Li, Huang, Yuan, He, Lin, He, He, and
  Yuan]{yan2025gptimgeval}
Zhiyuan Yan, Junyan Ye, Weijia Li, Zilong Huang, Shenghai Yuan, Xiangyang He,
  Kaiqing Lin, Jun He, Conghui He, and Li~Yuan.
\newblock {GPT-ImgEval}: A comprehensive benchmark for diagnosing {GPT4o} in
  image generation, 2025.

\bibitem[Yang et~al.(2023)Yang, Gui, Yuan, Liang, Ding, Hu, and
  Chen]{yang2023glyphcontrol}
Yukang Yang, Dongnan Gui, Yuhui Yuan, Weicong Liang, Haisong Ding, Han Hu, and
  Kai Chen.
\newblock {GlyphControl}: Glyph conditional controllable visual text
  generation.
\newblock In \emph{Thirty-seventh Conference on Neural Information Processing
  Systems}, 2023.

\bibitem[Zhao et~al.(2026)Zhao, Zhang, Tang, Zhu, Li, Chai, Zhang, Xia, Zhai,
  Yan, Yang, Yang, and Duan]{zhao2025risebench}
Xiangyu Zhao, Peiyuan Zhang, Kexian Tang, Xiaorong Zhu, Hao Li, Wenhao Chai,
  Zicheng Zhang, Renqiu Xia, Guangtao Zhai, Junchi Yan, Hua Yang, Xue Yang, and
  Haodong Duan.
\newblock Envisioning beyond the pixels: Benchmarking reasoning-informed visual
  editing.
\newblock In \emph{The Thirty-ninth Annual Conference on Neural Information
  Processing Systems Datasets and Benchmarks Track}, 2026.

\bibitem[Zhou et~al.(2024)Zhou, Yu, Babu, Tirumala, Yasunaga, Shamis, Kahn, Ma,
  Zettlemoyer, and Levy]{zhou2024transfusion}
Chunting Zhou, Lili Yu, Arun Babu, Kushal Tirumala, Michihiro Yasunaga, Leonid
  Shamis, Jacob Kahn, Xuezhe Ma, Luke Zettlemoyer, and Omer Levy.
\newblock Transfusion: Predict the next token and diffuse images with one
  multi-modal model, 2024.

\end{thebibliography}
